\documentclass{article}

\usepackage{arxiv}

\usepackage[utf8]{inputenc} 
\usepackage[T1]{fontenc}    
\usepackage{hyperref}       
\usepackage{url}            
\usepackage{booktabs}       
\usepackage{amsfonts}       
\usepackage{amsmath}
\usepackage{nicefrac}       
\usepackage{microtype}      
\usepackage{graphicx}
\usepackage{natbib}
\usepackage{doi}
\usepackage{enumitem}
\usepackage{adjustbox}
\usepackage{tabularx}
\usepackage{pdflscape}

\title{Advancing Polish Language Modeling through Tokenizer Optimization in the Bielik v3 7B and 11B Series}

\author{
 \textbf{Krzysztof Ociepa\textsuperscript{1,4}},
 \textbf{Łukasz Flis\textsuperscript{1,2}},
 \\
 \textbf{Remigiusz Kinas\textsuperscript{1}},
 \textbf{Krzysztof Wróbel\textsuperscript{1,3,5}},
 \textbf{Adrian Gwoździej\textsuperscript{1,2}}
\\
\\
 \textsuperscript{1}SpeakLeash,
 \textsuperscript{2}ACK Cyfronet AGH,
 \textsuperscript{3}Jagiellonian University,
 \textsuperscript{4}Azurro,
 \textsuperscript{5}Enelpol
}

\date{}

\renewcommand{\undertitle}{\includegraphics[height=40pt]{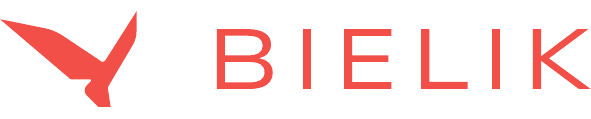}}
\renewcommand{\shorttitle}{\includegraphics[height=15pt]{./logo/bielik-logo.pdf}}

\hypersetup{
pdftitle={Bielik v3 PL},
pdfsubject={cs.CL},
pdfauthor={Krzysztof Ociepa, Łukasz Flis, Remigiusz Kinas, Krzysztof Wróbel, Adrian Gwoździej},
}

\begin{document}
\maketitle

\begin{abstract}
The development of the Bielik v3 PL series, encompassing both the 7B and 11B parameter variants, represents a significant milestone in the field of language-specific large language model (LLM) optimization. While general-purpose models often demonstrate impressive multilingual capabilities, they frequently suffer from a fundamental architectural inefficiency: the use of universal tokenizers. These tokenizers, typically designed to cover a broad spectrum of languages, often fail to capture the morphological nuances of specific languages like Polish, leading to higher fertility ratios, increased inference costs, and restricted effective context windows. This report details the transition from the universal Mistral-based tokenization to a dedicated Polish-optimized vocabulary for the Bielik v3 models, exploring the FOCUS-based embedding initialization, the multi-stage pretraining curriculum, and the subsequent post-training alignment involving Supervised Fine-Tuning, Direct Preference Optimization, and Reinforcement Learning through Group Relative Policy Optimization with verifiable rewards.
\end{abstract}


\section{Introduction}

Recent years have witnessed rapid progress in the development of large language models (LLMs), including a growing focus on languages that remain underrepresented in global AI systems. Within Europe, multiple initiatives have emerged to support linguistic diversity and improve access to high-quality language technologies across a wide range of languages. 

Our work extends the Bielik model family, building upon the Bielik 11B v3 model \cite{ociepa2025bielik11bv3multilingual} and the Bielik Minitron 7B v3 model \cite{kinas2026bielikminitron7bcompressinglargelanguage}. The approach presented in this paper leverages prior experience and methodologies developed during earlier iterations of smaller Bielik v3 models \cite{ociepa2025bielikv3smalltechnical}. 

This research is aligned with broader European efforts aimed at advancing multilingual and accessible AI systems. Notable examples include EuroLLM \cite{MARTINS202553}, which focuses on multilingual capabilities across European Union languages, Apertus \cite{swissai2025apertus}, which promotes open and inclusive LLM development, and the PLLuM family \cite{kocon2025pllumfamilypolishlarge}, which targets Polish language modeling specifically.

In this paper, we introduce Bielik v3 PL models in both 7B and 11B parameter configurations, designed with a tokenizer optimized specifically for the Polish language. The main contributions of this work are as follows:

\begin{itemize}
    \item We propose a method for replacing the tokenizer using the FOCUS \cite{dobler2023focus} approach, while mitigating the risk of catastrophic forgetting.
    \item We describe a comprehensive multi-stage pre-training and post-training pipeline that preserves performance comparable to models using the original tokenizer.
    \item We release the weights of both models under the Apache 2.0 license.
\end{itemize}

\section{Model Architecture}

The Bielik v3 family is based on the Transformer architecture \cite{Vaswani2017AttentionIA}, adopting and extending key design principles introduced in the Mistral 7B models \cite{jiang2023mistral7b}. The architecture incorporates several optimizations aimed at improving both computational efficiency and training robustness, while maintaining strong performance across a wide range of tasks.

A central component of the design is the use of Grouped-Query Attention (GQA) \cite{ainslie-etal-2023-gqa}, which reduces memory bandwidth usage and computational overhead during inference. This is achieved by sharing key-value projections across multiple query heads, effectively lowering the number of key-value heads without significantly impacting model quality. This approach has become a standard technique in modern efficient large-scale models, particularly for handling long input sequences.

To support extended context lengths, the models employ Rotary Positional Embeddings (RoPE) \cite{SU2024127063}, which provide improved generalization of positional information compared to traditional embedding methods. This enables the Bielik v3 models to operate with a native context window of up to 32,768 tokens, while preserving sensitivity to token order over long sequences.

The flagship Bielik 11B v3 model \cite{ociepa2025bielik11bv3multilingual} was scaled using the Depth Up-Scaling (DUS) strategy \cite{kim2024solar107bscalinglarge}. Starting from a 32-layer Mistral-based backbone, the architecture was expanded by duplicating layers, followed by a structured reduction in which selected layers from both the lower and upper parts of the network were removed. This process resulted in a 50-layer model, balancing increased representational capacity with practical deployment constraints. The final architecture was specifically designed to fit within the memory limits of widely available 24~GB GPUs, while providing sufficient depth for advanced reasoning capabilities.

The Bielik Minitron 7B v3 model \cite{kinas2026bielikminitron7bcompressinglargelanguage} was obtained through compression of the 11B variant rather than being trained independently. This process combined structured pruning with knowledge distillation, enabling a substantial reduction in model size and computational requirements. As a result, the compressed model retains approximately 90\% of the original model's performance, while achieving up to 50\% faster inference. This approach significantly lowers both development cost and environmental impact, and demonstrates an effective strategy for building high-quality models for underrepresented languages.

The Bielik v3 PL variants retain the same architectural design as their base models, differing only in the tokenizer and vocabulary, which are specifically adapted for the Polish language.

\section{Tokenizers}

Tokenization defines the boundary between raw text and its numerical representation, making it a critical component of any language model. Its design is particularly important for morphologically rich languages such as Polish, which exhibits complex inflectional patterns, frequent use of diacritics, and a high degree of lexical variation. In such settings, suboptimal tokenization can significantly degrade model efficiency and performance.

General-purpose tokenizers, including those used in models such as Mistral 7B, are typically optimized for multilingual coverage rather than language-specific efficiency. As a result, Polish text is often segmented into an excessive number of subword units. This behavior is commonly measured using the \textit{fertility ratio}, defined as the average number of tokens required to represent a given text \cite{rust2021goodtokenizermonolingualperformance}. A high fertility ratio leads to reduced information density within the context window and increased computational cost during inference.

At the opposite end of the design spectrum are tokenizers with very large vocabularies, often ranging from 150k to 250k tokens. While such approaches can reduce fragmentation, they introduce significant overhead in terms of model size and memory consumption. In monolingual or language-focused applications, a substantial portion of these embeddings remains unused, resulting in inefficient utilization of both memory and compute resources, as well as slower inference.

The original Bielik v3 tokenizer employed a vocabulary of 32,128 tokens. Although effective, it frequently required multiple tokens to encode single Polish words that could otherwise be represented more compactly. This limitation negatively impacts both context utilization and generation speed. To address this issue, the Bielik v3 PL models adopt a dedicated Polish tokenizer with a comparable vocabulary size of 32,000 tokens. The primary objective of this design is to reduce the fertility ratio for Polish while preserving reasonable coverage of English and other European languages.

Beyond vocabulary size, the segmentation strategy itself plays a crucial role. In particular, the handling of digits, punctuation, and special characters can influence both token efficiency and downstream generation quality. Taking these factors into account, we developed and adopted the APT4 tokenizer, which extends and refines the design of the earlier APT3 tokenizer introduced with the Polish APT3 model \cite{AzurroAPT3Base1B}.

To quantitatively evaluate tokenizer performance, we use the preamble of the Constitution of the Republic of Poland (see Appendix~\ref{app:preamble}) as a benchmark. This text provides a representative example of formal Polish, characterized by complex syntax and rich morphology, while its official English translation enables controlled cross-linguistic comparison. 

Table~\ref{tab:tokenizers-comparison} reports key efficiency metrics, including the total number of tokens, characters per token (CpT), and tokens per word (TpW), for both Polish and English versions of the text. These metrics offer a concise and interpretable measure of how effectively each tokenizer encodes linguistic information, highlighting trade-offs between vocabulary size, segmentation granularity, and cross-lingual performance. In general, higher CpT indicates denser tokenization, while lower TpW indicates fewer tokens per word.

\begin{table*}[h]
\centering
\begin{tabular}{lll|lll|lll}
\toprule
                    &                 &                & \multicolumn{3}{c|}{Polish} & \multicolumn{3}{c}{English} \\
Tokenizer           & Vocab Size & Avg tokens & Tokens   & CpT    & TpW    & Tokens    & CpT    & TpW    \\
\midrule
APT3                           & 31980           & 480            & 344      & 5.22   & 1.48   & 615       & 3.15   & 1.93   \\
APT4                           & 32000           & 503            & 375      & 4.78   & 1.62   & 631       & 3.07   & 1.98   \\
Bielik 11B v3                  & 32128           & 578            & 747      & 2.40   & 3.22   & 408       & 4.75   & 1.28   \\
EuroLLM                        & 128000          & 421            & 437      & 4.11   & 1.88   & 404       & 4.79   & 1.27   \\
Llama 3.2/SmolLM3              & 128256          & 512            & 653      & 2.75   & 2.81   & 371       & 5.22   & 1.17   \\
Apertus/Mistral Small 3.1 24B  & 131072          & 462            & 547      & 3.28   & 2.36   & 377       & 5.14   & 1.19   \\
Qwen3                          & 151669          & 499            & 625      & 2.87   & 2.69   & 373       & 5.19   & 1.17   \\
Gemma3                         & 262145          & 447            & 510      & 3.52   & 2.20   & 383       & 5.05   & 1.20   \\
\bottomrule
\end{tabular}
\caption{Comparison of token count, characters per token (CpT), and tokens per word (TpW) for the preamble of the Constitution of the Republic of Poland in Polish and English, processed by various tokenizers with different vocabulary sizes.}
\label{tab:tokenizers-comparison}
\end{table*}

We keep the vocabulary size at approximately 32k for the Bielik v3 PL tokenizer to isolate improvements from segmentation efficiency rather than increasing vocabulary capacity.

\section{Vocabulary Adaptation}

Replacing the tokenizer of a pretrained language model introduces a substantial risk of \textit{catastrophic forgetting}, where previously acquired semantic and syntactic knowledge is degraded during the transition to a new embedding space. To mitigate this issue, the Bielik v3 PL models adopt the FOCUS (Fast Overlapping Token Combinations Using Sparsemax) framework \cite{dobler2023focus}, which enables a structured transfer of knowledge between vocabularies.

The FOCUS method represents each token in the target vocabulary as a sparse linear combination of tokens from the original vocabulary, selected based on semantic similarity in an auxiliary embedding space. This approach preserves relationships encoded during pretraining while enabling efficient adaptation to a new tokenization scheme. Our choice of FOCUS is supported by prior experimental results on earlier Bielik v3 models \cite{ociepa2025bielikv3smalltechnical}, where multiple embedding initialization strategies were systematically evaluated.

The methods considered include:

\begin{itemize}
    \item \textbf{Random Initialization:} Assigns randomly sampled vectors to new tokens, requiring the model to relearn embeddings from scratch, often resulting in slow convergence.

    \item \textbf{Frequency-based Vocabulary Transfer (FVT)} \cite{yuan2022frequency}: Initializes token embeddings by aggregating representations of their constituent subword units, guided by frequency statistics.

    \item \textbf{Linear Transformation (aX + b):} Maps embeddings between vocabularies via a learned linear projection, aiming to preserve geometric structure.

    \item \textbf{WECHSEL} \cite{minixhofer2022wechsel}: Uses multilingual static embeddings to align semantically related tokens across vocabularies.

    \item \textbf{FOCUS} \cite{dobler2023focus}: Constructs token embeddings as sparse combinations of semantically overlapping tokens, improving precision and stability.

    \item \textbf{MATT (Model-Aware Tokenizer Transfer)} \cite{haltiuk2025modelawaretokenizertransfer}: Extends FOCUS by incorporating attention-based objectives that preserve inter-token interaction patterns.

    \item \textbf{OFA (One For All)} \cite{liu2023ofa}: Relies on external multilingual embeddings to initialize unseen tokens across languages.

    \item \textbf{RAMEN} \cite{tran2020from}: Applies cross-lingual alignment techniques, such as bilingual lexicons, to transfer embeddings between languages.
\end{itemize}

Among these approaches, FOCUS consistently demonstrated the best empirical performance. In particular, experiments conducted on the Bielik 1.5B v3 model showed the lowest training loss after 4B tokens of continued pretraining, as well as leading results on the Open Polish LLM Leaderboard \cite{open-pl-llm-leaderboard, ociepa2024bielik7bv01polish}. By leveraging Sparsemax for token selection, FOCUS restricts the combination to the most relevant components, resulting in high-quality initialization of the new embedding matrix. This leads to stable optimization behavior during subsequent training phases.

\subsection{Multi-Stage Continued Pretraining Pipeline}

\begin{figure}[h!]
\centering
\begin{minipage}{0.48\textwidth}
    \centering
    \includegraphics[width=\textwidth]{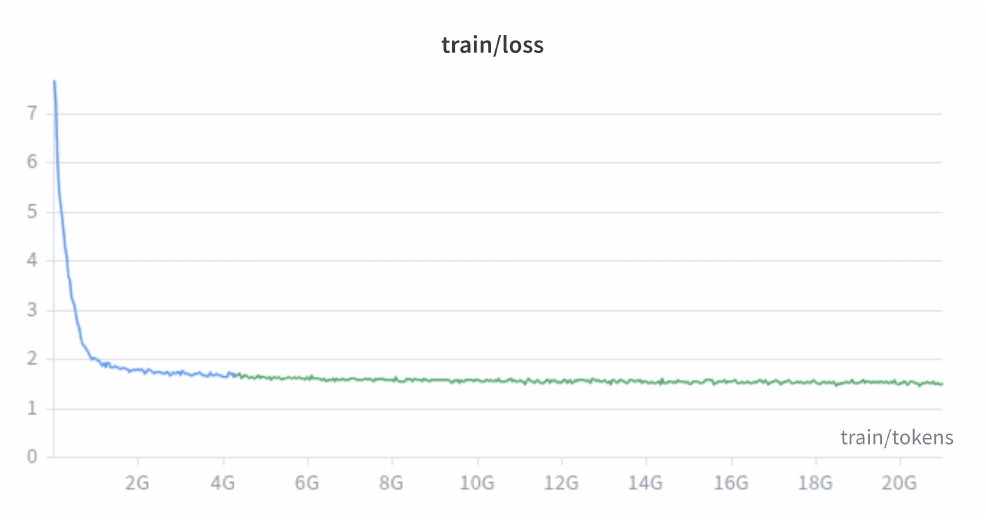}
    \caption{Training loss over the training tokens for the Bielik 11B v3 PL model.}
    \label{fig:base-training-loss}
\end{minipage}
\hfill
\begin{minipage}{0.48\textwidth}
    \centering
    \includegraphics[width=\textwidth]{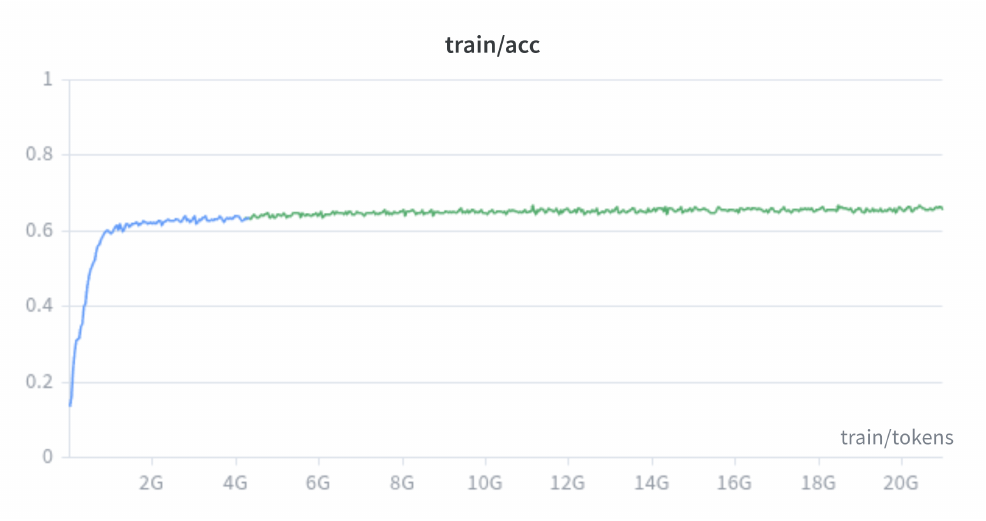}
    \caption{Training accuracy over the training tokens for the Bielik 11B v3 PL model.}
    \label{fig:base-training-acc}
\end{minipage}
\end{figure}

To adapt the model to the new tokenizer while preserving its internal representations, we employ a two-stage continued pretraining procedure. The training data consists of a 20B-token subset sampled from the original Bielik 11B v3 corpus, ensuring consistency in distribution and domain coverage. Training loss and accuracy over the training tokens for the Bielik 11B v3 PL model are presented in Figures \ref{fig:base-training-loss} and \ref{fig:base-training-acc}.

\subsubsection{Stage 1: Partial Freezing and Boundary Adaptation}

The first stage focuses on stabilizing the interaction between the new tokenizer and the pretrained model. Continued pretraining is performed on 4B tokens, while most of the model parameters remain frozen. Only the following components are updated:

\begin{itemize}
    \item The input embedding layer,
    \item The language modeling head (\texttt{lm\_head}),
    \item Four boundary transformer layers (two lowest and two highest layers).
\end{itemize}

This selective training strategy constrains the adaptation process to a limited subset of parameters, effectively learning a mapping between the new token space and the fixed internal representations. By restricting updates to boundary layers, the model preserves its higher-level reasoning capabilities while gradually aligning with the new vocabulary. Empirically, this phase is critical for ensuring training stability and preventing divergence during later stages.

\subsubsection{Stage 2: Full Model Adaptation}

After initial stabilization, all model parameters are unfrozen. The model then undergoes continued pretraining on an additional 16B tokens. This phase allows the network to globally adjust its weights, refining both linguistic representations and token-level statistics to better match the characteristics of the Polish language.

\subsection{Post-Training}

Following tokenizer adaptation and continued pretraining, the Bielik v3 PL models are subjected to the same post-training pipeline as the original Bielik v3 models \cite{ociepa2025bielik11bv3multilingual}. This ensures a fair and consistent comparison across model variants.

\begin{enumerate}
    \item \textbf{Supervised Fine-Tuning (SFT):} The model is first fine-tuned on a curated dataset of high-quality instruction-response pairs in Polish and English. This stage establishes the model's conversational abilities and aligns it with expected formatting and linguistic norms. Training is conducted for 3 epochs on approximately 20 million samples, with a maximum sequence length of 32{,}768 tokens.

    \item \textbf{Preference Optimization (DPO-P)} \cite{pal2024smaugfixingfailuremodes}: We apply Direct Preference Optimization in its positive-only formulation, which emphasizes stable policy improvement while reinforcing desirable outputs. This stage reduces hallucinations and improves adherence to user intent. Training is performed for 3 epochs on a dataset of 114{,}000 preference-labeled examples.

    \item \textbf{Reinforcement Learning (GRPO)} \cite{shao2024deepseek}: To enhance reasoning capabilities, we incorporate Group Relative Policy Optimization. Using verifiable reward signals in domains such as mathematics, logic, and STEM tasks, this stage enables iterative refinement of intermediate reasoning steps without requiring an explicit critic model. The training set consists of 143{,}000 specialized examples.
\end{enumerate}

\section{Evaluation}
\label{sec:evaluation}

The critical success criterion for the Bielik v3 PL models was to maintain the benchmark performance of the source models while achieving the aforementioned token efficiency. We report \textbf{Bielik-PL-11B-v3.0-Instruct} and \textbf{Bielik-PL-Minitron-7B-v3.0-Instruct} (checkpoints with the Polish tokenizer) alongside the full leaderboard comparisons from the Bielik 11B v3 technical report \cite{ociepa2025bielik11bv3multilingual}.

To comprehensively assess the capabilities of Bielik v3 models, we conducted extensive evaluations across multiple benchmarks covering diverse aspects of language understanding, generation, and reasoning. Our evaluation strategy encompasses both Polish-specific and multilingual benchmarks to demonstrate the models' proficiency in handling various linguistic tasks.

The models were evaluated on the following benchmarks:

\begin{itemize}
    \item \href{https://huggingface.co/spaces/speakleash/open_pl_llm_leaderboard}{Open PL LLM Leaderboard} \textbf{(Polish)}
    \item \href{https://huggingface.co/spaces/speakleash/polish_eq-bench}{Polish EQ-Bench} \textbf{(Polish)}
    \item \href{https://huggingface.co/spaces/speakleash/cptu_bench}{CPTUB Leaderboard} \textbf{(Polish)}
    \item \href{https://huggingface.co/spaces/speakleash/polish_medical_leaderboard}{Polish Medical Leaderboard} \textbf{(Polish)}
    \item \href{https://huggingface.co/spaces/sdadas/plcc}{Polish Linguistic and Cultural Competency Benchmark (PLCC)} \textbf{(Polish)}
    \item \href{https://huggingface.co/spaces/open-llm-leaderboard-old/open_llm_leaderboard}{Open LLM Leaderboard}
    \item \href{https://huggingface.co/spaces/speakleash/include-base-european-leaderboard}{include-base-44}
    \item \href{https://huggingface.co/spaces/speakleash/belebele-european-leaderboard}{belebele}
    \item \href{https://huggingface.co/spaces/speakleash/leaderboard-flores}{flores}
\end{itemize}

\subsection{Open PL LLM Leaderboard} \label{Open-PL-LLM-Leaderboard}
The Open PL LLM Leaderboard \cite{open-pl-llm-leaderboard, ociepa2024bielik7bv01polish} provides a comprehensive assessment of language models across a diverse range of Polish NLP tasks. Built upon the foundation of Open LLM Leaderboard v1 \cite{open-llm-leaderboard-v1}, this benchmark evaluates core language understanding capabilities including sentiment classification, named entity recognition, topic categorization, reading comprehension, and question answering. The evaluation framework employs the lm-evaluation-harness toolkit \cite{eval-harness} and primarily focuses on discrete task performance rather than conversational interaction abilities.

\paragraph{Tasks:}
\begin{itemize}
    \item \textbf{polemo2:} Sentiment analysis of online consumer reviews across four domains (medicine, hotels, products, university) with four-class labeling (positive, negative, neutral, ambiguous) \cite{kocon-etal-2019-multi}; metric: accuracy.
    \item \textbf{klej-ner:} Named entity recognition in sentences containing single-type entities, classifying into six categories (no entity, place, person, organization, time, geographical name) \cite{rybak-etal-2020-klej}; metric: accuracy.
    \item \textbf{8tags:} Topic classification of social media headlines into eight categories (film, history, food, medicine, motorization, work, sport, technology) \cite{dadas-etal-2020-evaluation}; metric: accuracy.
    \item \textbf{belebele:} Machine reading comprehension for question answering \cite{bandarkar-etal-2024-belebele}; metric: accuracy (as used within the Open PL LLM Leaderboard task suite; see Section~\ref{sec:evaluation} for separate Belebele subset reporting).
    \item \textbf{dyk:} Question answering based on human-annotated pairs from Wikipedia's "Did You Know" section \cite{marcinczuk2013open}; metric: binary F1.
    \item \textbf{ppc:} Text similarity assessment using manually labeled sentence pairs (exact paraphrases, close paraphrases, non-paraphrases) \cite{9945218}; metric: accuracy.
    \item \textbf{psc:} Summarization of news articles \cite{ogro:kop:14:lrec}; metric: binary F1.
    \item \textbf{cbd:} Text classification for cyberbullying and hate-speech detection \cite{ptaszynski2023expert}; metric: macro F1.
    \item \textbf{polqa:} Open-domain question answering from the "Jeden z dziesi\k{e}ciu" TV show, with and without context (abstractive QA/RAG) \cite{rybak-etal-2024-polqa-polish}; metric: accuracy, levenshtein.
    \item \textbf{poquad:} Context-based extractive question answering (QA/RAG) \cite{tuora2023poquad}; metric: levenshtein.
    \item \textbf{eqbench:} emotional intelligence benchmark \cite{paech2024eqbenchemotionalintelligencebenchmark}; metric: custom.
\end{itemize}

The majority of benchmark tasks employ a multiple-choice format where models select from predefined answer options. Two distinct evaluation methodologies are applied:
\begin{itemize}
    \item \textbf{Loglikelihood:} Models select the option with the highest token probability from the available choices (e.g., A, B, C, D). This approach is particularly well-suited for evaluating base models without instruction tuning.
    \item \textbf{Generate:} Models produce free-form text responses, testing their generation capabilities.
\end{itemize}

Each task undergoes evaluation in both 0-shot (no examples provided) and 5-shot (five examples given) configurations, with final scores normalized against a random-choice baseline for the given number of answer options. Table~\ref{tab:open-pl-llm-instruct} reports 5-shot averages for instruction-tuned models, including Bielik-11B-v3.0-Instruct and the Bielik-PL variants.

  \begin{table*}[t]
    \centering
    \small
    \begin{tabular}{lrr}
    \toprule
    \textbf{Model} & \textbf{Parameters (B)} & \textbf{Average} \\
    \midrule
    Mistral-Large-Instruct-2411 & 123.0 & 69.84 \\
    Meta-Llama-3.1-405B-Instruct-FP8 & 405.0 & 69.44 \\
    Mistral-Large-Instruct-2407 & 123.0 & 69.11 \\
    Qwen2.5-72B-Instruct & 72.7 & 67.92 \\
    QwQ-32B-Preview & 32.8 & 67.01 \\
    Llama-3.3-70B-Instruct & 70.6 & 66.40 \\
    \textbf{Bielik-11B-v3.0-Instruct} & \textbf{11.2} & \textbf{65.93} \\
    Qwen2-72B-Instruct & 72.7 & 65.87 \\
    \underline{Bielik-11B-v2.3-Instruct} & \underline{11.2} & \underline{65.71} \\
    \underline{Bielik-11B-v2.2-Instruct} & \underline{11.2} & \underline{65.57} \\
    Meta-Llama-3.1-70B-Instruct & 70.6 & 65.49 \\
    \underline{Bielik-11B-v2.1-Instruct} & \underline{11.2} & \underline{65.45} \\
    Mixtral-8x22B-Instruct-v0.1 & 141.0 & 65.23 \\
    \underline{Bielik-11B-v2.0-Instruct} & \underline{11.2} & \underline{64.98} \\
    Meta-Llama-3-70B-Instruct & 70.6 & 64.45 \\
    \underline{Bielik-11B-v2.6-Instruct} & \underline{11.2} & \underline{64.26} \\
    Qwen3-32B & 32.8 & 64.24 \\
    Llama-4-Scout-17B-16E-Instruct & 109.0 & 64.21 \\
    \textbf{Bielik-PL-11B-v3.0-Instruct} & \textbf{11.2} & \textbf{64.11} \\
    \underline{Bielik-11B-v2.5-Instruct} & \underline{11.2} & \underline{63.95} \\
    Mistral-Small-24B-Instruct-2501 & 24.0 & 62.97 \\
    phi-4 & 14.7 & 62.57 \\
    \textbf{Bielik-Minitron-7B-v3.0-Instruct} & \textbf{7.5} & \textbf{62.46} \\
    Qwen3-14B & 14.8 & 62.24 \\
    gemma-3-12b-it & 12.0 & 62.20 \\
    \textbf{Bielik-PL-Minitron-7B-v3.0-Instruct} & \textbf{7.5} & \textbf{61.66} \\
    Mistral-Small-Instruct-2409 & 22.2 & 61.41 \\
    Qwen2.5-32B-Instruct & 32.8 & 61.21 \\
    Qwen2.5-14B-Instruct & 14.8 & 59.91 \\
    aya-23-35B & 35.0 & 56.37 \\
    \underline{Bielik-4.5B-v3.0-Instruct} & \underline{4.8} & \underline{56.13} \\
    gemma-3-27b-it & 27.0 & 55.92 \\
    Qwen3-8B & 8.2 & 55.78 \\
    Qwen3-4B & 4.0 & 55.49 \\
    Mistral-Nemo-Instruct-2407 & 12.2 & 55.27 \\
    EuroLLM-22B-Instruct-Preview & 22.0 & 55.17 \\
    Qwen2.5-7B-Instruct & 7.6 & 54.93 \\
    EuroLLM-9B-Instruct & 9.2 & 50.07 \\
    GaMS-9B-Instruct & 9.0 & 48.78 \\
    Mistral-7B-Instruct-v0.3 & 7.2 & 47.74 \\
    Apertus-8B-Instruct-2509 & 8.0 & 47.27 \\
    Mistral-7B-Instruct-v0.2 & 7.2 & 45.95 \\
    \underline{Bielik-7B-Instruct-v0.1} & \underline{7.2} & \underline{44.70} \\
    gemma-2-9b-it & 9.0 & 42.12 \\  
    Qwen2.5-3B-Instruct & 3.0 & 41.23 \\
    Mistral-7B-Instruct-v0.1 & 7.0 & 33.11 \\
    Qwen2.5-1.5B-Instruct & 1.5 & 31.89 \\
    \bottomrule
    \end{tabular}
    \caption{Open PL LLM Leaderboard results for instruction-tuned models (5-shot evaluation)}
    \label{tab:open-pl-llm-instruct}
    \end{table*}

As shown in Table~\ref{tab:open-pl-llm-instruct}, Bielik-11B-v3.0-Instruct achieves a 5-shot average of 65.93, ranking among the top models listed and outperforming several much larger models in the same table, including Meta-Llama-3.1-70B-Instruct and Mixtral-8x22B-Instruct-v0.1. The Polish tokenizer checkpoints, Bielik-PL-11B-v3.0-Instruct and Bielik-PL-Minitron-7B-v3.0-Instruct, achieve 64.11 and 61.66 on the same 5-shot aggregate. For reference, \textbf{Bielik-Minitron-7B-v3.0-Instruct} (compressed 7B with the original tokenizer) scores 62.46 \cite{kinas2026bielikminitron7bcompressinglargelanguage}.

    \subsection{Polish EQ-Bench}

    The Polish Emotional Intelligence Benchmark represents a culturally adapted Polish adaptation of the EQ-Bench framework \cite{paech2024eqbenchemotionalintelligencebenchmark}. This benchmark assesses language models' ability to recognize, interpret, and reason about emotional states and interpersonal dynamics. The evaluation encompasses multiple facets of emotional intelligence, including emotion recognition in context, understanding of emotional implications, and sensitivity to nuanced affective states in conversational scenarios. Results are presented in Table~\ref{tab:pl-eq-bench}.
    
    \begin{table*}[t]
    \centering
    \small
    \begin{tabular}{lrc}
    \toprule
    \textbf{Model} & \textbf{Parameters (B)} & \textbf{Score} \\
    \midrule
    Mistral-Large-Instruct-2407$^{\dagger}$ & 123.0 & 78.07 \\
    Mistral-Large-Instruct-2411$^{\dagger}$ & 123.0 & 77.29 \\
    Meta-Llama-3.1-405B-Instruct-FP8 & 405.0 & 77.23 \\
    gpt-4o-2024-08-06 & Unknown & 75.15 \\
    gpt-4-turbo-2024-04-09 & Unknown & 74.59 \\
    \underline{Bielik-11B-v2.6-Instruct} & \underline{11.2} & \underline{73.8} \\
    DeepSeek-V3-0324 & 685.0 & 73.46 \\
    Mistral-Small-Instruct-2409 & 22.2 & 72.85 \\
    Llama-PLLuM-70B-chat & 70.6 & 72.56 \\
    Meta-Llama-3.1-70B-Instruct & 70.6 & 72.53 \\
    \underline{Bielik-11B-v2.5-Instruct} & \underline{11.2} & \underline{72.00} \\
    Qwen2-72B-Instruct & 72.7 & 71.23 \\
    Meta-Llama-3-70B-Instruct & 70.6 & 71.21 \\
    \textbf{Bielik-11B-v3.0-Instruct} & \textbf{11.2} & \textbf{71.20} \\
    \textbf{Bielik-PL-11B-v3.0-Instruct} & \textbf{11.2} & \textbf{71.15} \\
    gpt-4o-mini-2024-07-18 & Unknown & 71.15 \\
    Qwen2.5-32B-Instruct & 32.8 & 71.15 \\
    \underline{Bielik-11B-v2.3-Instruct} & \underline{11.2} & \underline{70.86} \\
    Llama-3.3-70B-Instruct & 70.6 & 70.73 \\
    Llama-PLLuM-70B-instruct & 70.6 & 69.99 \\
    WizardLM-2-8x22B & 141.0 & 69.56 \\
    Qwen2.5-14B-Instruct & 14.8 & 69.17 \\
    \underline{Bielik-11B-v2.2-Instruct} & \underline{11.2} & \underline{69.05} \\
    \underline{Bielik-11B-v2.0-Instruct} & \underline{11.2} & \underline{68.24} \\
    \textbf{Bielik-PL-Minitron-7B-v3.0-Instruct} & \textbf{7.5} & \textbf{66.89} \\
    \textbf{Bielik-Minitron-7B-v3.0-Instruct} & \textbf{7.5} & \textbf{64.09} \\
    glm-4-9b-chat & 9.0 & 61.79 \\
    Mistral-Nemo-Instruct-2407 & 12.2 & 61.76 \\
    \underline{Bielik-11B-v2.1-Instruct} & \underline{11.2} & \underline{60.07} \\
    pllum-12b-nc-chat-250715 & 12.2 & 55.20 \\
    EuroLLM-9B-Instruct & 9.2 & 54.10 \\
    \underline{Bielik-4.5B-v3.0-Instruct} & \underline{4.8} & \underline{53.58} \\
    PLLuM-12B-chat & 12.2 & 52.26 \\
    PLLuM-8x7B-nc-chat$^{\dagger}$ & 46.7 & 47.29 \\
    Llama-PLLuM-8B-chat & 8.0 & 46.20 \\
    PLLuM-8x7B-chat & 46.7 & 45.22 \\
    PLLuM-12B-nc-chat$^{\dagger}$ & 12.2 & 35.41 \\
    \bottomrule
    \multicolumn{3}{l}{$^{\dagger}$Models with a non-commercial license.} \\
    \end{tabular}
    \caption{Polish EQ-Bench results for various models.}
    \label{tab:pl-eq-bench}
    \end{table*}

Bielik-11B-v3.0-Instruct achieves a score of 71.20 on the Polish EQ-Bench (Table~\ref{tab:pl-eq-bench}), demonstrating strong emotional intelligence capabilities. While this represents a slight decrease compared to the previous version Bielik-11B-v2.6-Instruct (73.8), the v3.0 model maintains competitive performance, placing it among models with substantially larger parameter counts such as Llama-3.3-70B-Instruct (70.73) and Qwen2.5-32B-Instruct (71.15). The Polish tokenizer variants, Bielik-PL-11B-v3.0-Instruct and Bielik-PL-Minitron-7B-v3.0-Instruct, score 71.15 and 66.89 on the \texttt{eq-bench\_v2\_pl} run reported in the same table. \textbf{Bielik-Minitron-7B-v3.0-Instruct} (original tokenizer) scores 64.09 \cite{kinas2026bielikminitron7bcompressinglargelanguage}.

    \subsection{Complex Polish Text Understanding Benchmark (CPTUB)}

    CPTUB \cite{cptub-leaderboard} presents a sophisticated evaluation framework targeting advanced comprehension capabilities in Polish language processing. In contrast to conventional benchmarks that primarily test literal interpretation, CPTUB probes deeper cognitive abilities including inference from context, pragmatic understanding, and reasoning under ambiguity. The benchmark structure encompasses two primary evaluation dimensions:
    
    \begin{itemize}
        \item \textbf{Implicatures}: This component measures models' competence in decoding non-literal meanings and contextual implications. It examines understanding of figurative language, ironic expressions, and idiomatic constructions through three distinct evaluation categories:
        \begin{itemize}
            \item \textbf{Sentiment}: Assessing the ability to discern emotional valence that diverges from surface-level lexical content
            \item \textbf{Language understanding}: Testing comprehension of communicative intent and pragmatic meaning
            \item \textbf{Phraseology}: Evaluating knowledge of conventionalized multi-word expressions where compositional semantics fail
        \end{itemize}
        \item \textbf{Tricky Questions}: This section challenges models with adversarially constructed queries featuring logical paradoxes, semantic ill-formedness, contradictions, absurdist premises, and humorous misdirection. It specifically probes reasoning robustness and the model's tendency to produce plausible-sounding but incorrect responses when confronted with problematic inputs.
    \end{itemize}
    
    Table~\ref{tab:cptub} presents the comprehensive results across all CPTUB evaluation categories.
    
    \begin{landscape}
    \begin{table*}[t]
    \centering
    \small
    \begin{tabular}{lrccccccc}
    \toprule
    \textbf{Model} & \textbf{Params (B)} & \textbf{Overall} & \textbf{Implicatures} & \textbf{Senti-} & \textbf{Language} & \textbf{Phrase-} & \textbf{Tricky} \\
    & & \textbf{Average} & \textbf{Average} & \textbf{ment} & \textbf{Understanding} & \textbf{ology} & \textbf{Questions} \\
    \midrule
    gemini-2.0-flash-001 & Unknown & 4.29 & 4.39 & 4.52 & 4.32 & 4.34 & 3.99 \\
    DeepSeek-R1 & 685.0 & 4.14 & 4.14 & 4.49 & 4.35 & 3.60 & 4.12 \\
    gemini-2.0-flash-lite-001 & Unknown & 4.09 & 4.17 & 4.23 & 4.05 & 4.24 & 3.85 \\
    DeepSeek-V3-0324 & 685.0 & 4.03 & 4.03 & 4.36 & 4.20 & 3.54 & 4.02 \\
    Mistral-Large-Instruct-2411$^{\dagger}$ & 123.0 & 4.00 & 4.10 & 4.33 & 3.98 & 3.99 & 3.72 \\
    Qwen2.5-72B-Instruct & 72.7 & 3.95 & 3.99 & 4.08 & 3.97 & 3.93 & 3.81 \\
    Mistral-Large-Instruct-2407$^{\dagger}$ & 123.0 & 3.93 & 4.03 & 4.23 & 4.00 & 3.86 & 3.65 \\
    Llama-4-Maverick-17B-128E-Instruct & 402.0 & 3.93 & 3.99 & 4.39 & 4.11 & 3.48 & 3.76 \\
    gemma-3-27b-it & 27.4 & 3.81 & 3.90 & 3.88 & 3.79 & 4.03 & 3.53 \\
    \textbf{Bielik-PL-11B-v3.0-Instruct} & \textbf{11.2} & \textbf{3.80} & \textbf{4.02} & \textbf{4.05} & \textbf{4.03} & \textbf{3.98} & \textbf{3.13} \\
    Meta-Llama-3-70B-Instruct & 70.6 & 3.78 & 3.81 & 4.13 & 3.82 & 3.47 & 3.71 \\
    Qwen2.5-32B-Instruct & 32.8 & 3.75 & 3.80 & 3.81 & 3.57 & 4.04 & 3.59 \\
    Llama-4-Scout-17B-16E-Instruct & 109.0 & 3.75 & 3.94 & 4.10 & 3.81 & 3.90 & 3.19 \\
    \textbf{Bielik-11B-v3.0-Instruct} & \textbf{11.2} & \textbf{3.73} & \textbf{3.92} & \textbf{3.88} & \textbf{3.91} & \textbf{3.96} & \textbf{3.19} \\
    Mistral-Small-24B-Instruct-2501 & 23.6 & 3.71 & 3.80 & 3.91 & 3.60 & 3.88 & 3.45 \\
    pllum-12b-nc-chat-250715$^{\dagger}$ & 12.2 & 3.67 & 3.92 & 4.36 & 3.96 & 3.46 & 2.90 \\
    \underline{Bielik-11B-v2.6-Instruct} & \underline{11.2} & \underline{3.64} & \underline{3.82} & \underline{4.10} & \underline{3.94} & \underline{3.41} & \underline{3.10} \\
    Mixtral-8x22B-Instruct-v0.1 & 141.0 & 3.56 & 3.67 & 3.78 & 3.68 & 3.55 & 3.24 \\
    Qwen2.5-14B-Instruct & 14.8 & 3.55 & 3.62 & 3.91 & 3.57 & 3.37 & 3.34 \\
    \textbf{Bielik-PL-Minitron-7B-v3.0-Instruct} & \textbf{7.5} & \textbf{3.55} & \textbf{3.87} & \textbf{3.88} & \textbf{3.82} & \textbf{3.92} & \textbf{2.58} \\
    Llama-PLLuM-70B-chat & 70.6 & 3.53 & 3.63 & 3.94 & 3.61 & 3.35 & 3.21 \\
    \underline{Bielik-4.5B-v3.0-Instruct} & \underline{4.8} & \underline{3.38} & \underline{3.68} & \underline{3.76} & \underline{3.61} & \underline{3.67} & \underline{2.46} \\
    \textbf{Bielik-Minitron-7B-v3.0-Instruct} & \textbf{7.5} & \textbf{3.38} & \textbf{3.59} & \textbf{3.72} & \textbf{3.83} & \textbf{3.23} & \textbf{2.74} \\
    phi-4 & 14.7 & 3.30 & 3.50 & 3.72 & 3.54 & 3.24 & 2.72 \\
    PLLuM-12B-chat & 12.2 & 3.14 & 3.32 & 3.32 & 3.21 & 3.43 & 2.59 \\
    PLLuM-8x7B-nc-instruct$^{\dagger}$ & 46.7 & 3.11 & 3.56 & 3.88 & 3.59 & 3.22 & 1.76 \\
    EuroLLM-9B-Instruct & 9.2 & 3.15 & 3.28 & 3.37 & 3.30 & 3.17 & 2.75 \\
    Qwen2.5-7B-Instruct & 7.6 & 3.07 & 3.23 & 3.56 & 3.03 & 3.10 & 2.58 \\
    PLLuM-8x7B-nc-chat$^{\dagger}$ & 46.7 & 3.03 & 3.44 & 3.76 & 3.48 & 3.08 & 1.80 \\
    Meta-Llama-3.1-8B-Instruct & 8.0 & 3.01 & 3.31 & 3.97 & 3.38 & 2.58 & 2.11 \\
    PLLuM-8x7B-chat & 46.7 & 3.01 & 3.41 & 3.44 & 3.45 & 3.35 & 1.78 \\
    Meta-Llama-3-8B-Instruct & 8.0 & 3.00 & 3.17 & 3.33 & 3.15 & 3.04 & 2.48 \\
    Llama-PLLuM-8B-chat & 8.0 & 2.92 & 3.14 & 3.13 & 2.93 & 3.36 & 2.25 \\
    \underline{Bielik-7B-Instruct-v0.1} & \underline{7.2} & \underline{2.88} & \underline{3.13} & \underline{3.59} & \underline{3.48} & \underline{2.32} & \underline{2.16} \\
    \bottomrule
    \multicolumn{8}{l}{$^{\dagger}$Models with a non-commercial license.} \\
    \end{tabular}
    \caption{Complex Polish Text Understanding Benchmark (CPTUB) results across different evaluation categories}
    \label{tab:cptub}
    \end{table*}
\end{landscape}

On CPTUB (Table~\ref{tab:cptub}), Bielik-11B-v3.0-Instruct achieves an overall average of 3.73, ranking competitively among models evaluated. The model performs particularly well on implicature understanding with an average of 3.92, demonstrating strong capabilities in language understanding (3.91), sentiment analysis (3.88), and phraseology (3.96). The tricky questions component yields a score of 3.19, reflecting the challenging nature of these adversarial queries. This performance places Bielik-11B-v3.0-Instruct ahead of several larger models including Qwen2.5-14B-Instruct and Mixtral-8x22B-Instruct-v0.1, while approaching the performance of frontier models with significantly higher parameter counts. Scores for \textbf{Bielik-Minitron-7B-v3.0-Instruct} are taken from the Minitron technical report \cite{kinas2026bielikminitron7bcompressinglargelanguage}. The Polish tokenizer checkpoints \textbf{Bielik-PL-11B-v3.0-Instruct} and \textbf{Bielik-PL-Minitron-7B-v3.0-Instruct} achieve overall averages of 3.80 and 3.55, respectively, with the 11B PL variant scoring above the original-tokenizer Bielik-11B-v3.0-Instruct (3.73) on this aggregate.

    \subsection{Polish Medical Leaderboard}

    The Polish Medical Leaderboard provides a domain-specific assessment of language models using authentic questions from the Polish State Specialization Examination (Pa\'{n}stwowy Egzamin Specjalizacyjny, PES) spanning 2018-2022. This benchmark measures both medical domain knowledge and clinical reasoning abilities within the Polish healthcare context. The evaluation employs the speakleash/PES-2018-2022 dataset, derived from amu-cai/PES-2018-2022 \cite{pokrywka2024gpt4passes}, and tests models' capacity to apply medical knowledge in scenarios similar to those encountered by medical professionals seeking board certification. Results are shown in Table~\ref{tab:medical-leaderboard}.
    
    \begin{table*}[t]
    \centering
    \small
    \begin{tabular}{lrc}
    \toprule
    \textbf{Model} & \textbf{Parameters (B)} & \textbf{Average (\%)} \\
    \midrule
    Meta-Llama-3.1-405B-Instruct-FP8 & 405.0 & 69.20 \\
    Mistral-Large-Instruct-2407$^{\dagger}$ & 123.0 & 64.28 \\
    Qwen2.5-72B-Instruct & 72.7 & 63.89 \\
    Meta-Llama-3.1-70B-Instruct & 70.6 & 61.75 \\
    Qwen2-72B-Instruct & 72.7 & 61.35 \\
    Meta-Llama-3-70B-Instruct & 70.6 & 57.51 \\
    Qwen2.5-32B & 32.8 & 55.69 \\
    Qwen2.5-32B-Instruct & 32.8 & 54.52 \\
    \textbf{Bielik-11B-v3.0-Instruct} & \textbf{11.2} & \textbf{50.21} \\
    Qwen2.5-14B-Instruct & 14.8 & 49.60 \\
    \textbf{Bielik-PL-11B-v3.0-Instruct} & \textbf{11.2} & \textbf{48.42} \\
    \textbf{Bielik-11B-v3-Base-20250730} & \textbf{11.2} & \textbf{45.86} \\
    \underline{Bielik-11B-v2.6-Instruct} & \underline{11.2} & \underline{44.88} \\
    \underline{Bielik-11B-v2.5-Instruct} & \underline{11.2} & \underline{44.85} \\
    GLM-4-9b-chat & 9.0 & 44.54 \\
    \textbf{Bielik-Minitron-7B-v3.0-Instruct} & \textbf{7.5} & \textbf{44.36} \\
    Mistral-Small-Instruct-2409 & 22.2 & 43.60 \\
    \underline{Bielik-4.5B-v3.0-Instruct} & \underline{4.8} & \underline{43.55} \\
    \textbf{Bielik-PL-Minitron-7B-v3.0-Instruct} & \textbf{7.5} & \textbf{43.35} \\
    \underline{Bielik-11B-v2.3-Instruct} & \underline{11.2} & \underline{43.26} \\
    \underline{Bielik-11B-v2.1-Instruct} & \underline{11.2} & \underline{43.16} \\
    \underline{Bielik-11B-v2.2-Instruct} & \underline{11.2} & \underline{43.05} \\
    Qwen2.5-7B-Instruct & 7.6 & 42.69 \\
    \underline{Bielik-11B-v2.0-Instruct} & \underline{11.2} & \underline{41.53} \\
    Meta-Llama-3.1-8B-Instruct & 8.0 & 40.60 \\
    Mistral-Nemo-Instruct-2407 & 12.2 & 40.36 \\
    \underline{Bielik-11B-v2} & \underline{11.2} & \underline{39.98} \\
    PLLuM-12B-nc-chat-250715$^{\dagger}$ & 12.2 & 38.53 \\
    PLLuM-12B-chat & 12.2 & 36.51 \\
    EuroLLM-9B-Instruct & 9.2 & 35.96 \\
    Mistral-7B-Instruct-v0.3 & 7.0 & 31.24 \\
    \underline{Bielik-7B-Instruct-v0.1} & \underline{7.2} & \underline{29.74} \\
    \bottomrule
    \multicolumn{3}{l}{$^{\dagger}$Models with a non-commercial license.} \\
    \end{tabular}
    \caption{Polish Medical Leaderboard results (5-shot setting) showing model performance on Polish Board Certification Examinations.}
    \label{tab:medical-leaderboard}
    \end{table*}

On the Polish Medical Leaderboard (Table~\ref{tab:medical-leaderboard}), Bielik-11B-v3.0-Instruct achieves 50.21\%, demonstrating substantial medical knowledge and clinical reasoning capabilities. This represents a significant improvement over the base model Bielik-11B-v3-Base-20250730 (45.86\%), highlighting the effectiveness of instruction tuning for specialized domain tasks. \textbf{Bielik-Minitron-7B-v3.0-Instruct} reaches 44.36\% \cite{kinas2026bielikminitron7bcompressinglargelanguage}. These results demonstrate Bielik's capability to handle domain-specific knowledge in the medical field when evaluated in Polish.

    \subsection{Open LLM Leaderboard}

    The Open LLM Leaderboard \cite{open-llm-leaderboard} serves as a comprehensive English-language evaluation suite, assessing models across diverse tasks including commonsense reasoning (ARC challenge, HellaSwag, WinoGrande), factual accuracy (TruthfulQA), broad knowledge (MMLU), and mathematical reasoning (GSM8K). This benchmark provides crucial insights into multilingual models' English language capabilities, which is particularly important for European models like Bielik that aim to balance strong native language performance with English proficiency. Table~\ref{tab:open-llm-instruct} presents results for selected instruction-tuned models.
 
    \begin{landscape}
    \begin{table*}[t]
    \centering
    \small
    \begin{tabular}{lccccccc}
    \toprule
    \textbf{Model} & \textbf{AVG} & \textbf{arc\_challenge} & \textbf{hellaswag} & \textbf{truthfulqa\_mc2} & \textbf{mmlu} & \textbf{winogrande} & \textbf{gsm8k} \\
    \midrule
    SOLAR-10.7B-Instruct-v1.0 & 74.20 & 71.08 & 88.16 & 71.43 & 66.21 & 83.58 & 64.75 \\
    Phi-3-medium-4k-instruct & 73.45 & 67.32 & 85.76 & 57.71 & 77.83 & 72.69 & 79.38 \\
    \textbf{Bielik-11B-v3.0-Instruct} & \textbf{72.45} & \textbf{64.59} & \textbf{81.96} & \textbf{54.25} & \textbf{71.11} & \textbf{77.19} & \textbf{85.60} \\
    \textbf{Bielik-PL-11B-v3.0-Instruct} & \textbf{71.49} & \textbf{64.68} & \textbf{81.31} & \textbf{54.69} & \textbf{70.99} & \textbf{76.32} & \textbf{80.97} \\
    \underline{Bielik-11B-v2.5-Instruct} & \underline{71.42} & \underline{61.95} & \underline{80.71} & \underline{53.17} & \underline{67.44} & \underline{79.72}	& \underline{85.52} \\
    \underline{Bielik-11B-v2.6-Instruct} & \underline{71.10} & 62.54 & 80.56 & 53.43 & 67.53 & 78.77 & 83.78 \\
    Bielik-11B-v2.2-Instruct & 69.86 & 59.90 & 80.16 & 58.34 & 64.34 & 75.30 & 81.12 \\
    \underline{Bielik-11B-v2.3-Instruct} & \underline{69.82} & 59.30 & 80.11 & 57.42 & 64.57 & 76.24 & 81.27 \\
    Bielik-11B-v2.1-Instruct & 69.82 & 59.56 & 80.20 & 59.35 & 64.18 & 75.06 & 80.59 \\
    openchat-3.5-0106-gemma & 69.42 & 64.68 & 81.08 & 54.93 & 64.69 & 78.30 & 72.86 \\
    Bielik-11B-v2.0-Instruct & 68.04 & 58.62 & 78.65 & 54.65 & 63.71 & 76.32 & 76.27 \\
    \textbf{Bielik-PL-Minitron-7B-v3.0-Instruct} & \textbf{67.63} & \textbf{57.51} & \textbf{74.80} & \textbf{53.40} & \textbf{65.31} & \textbf{73.16} & \textbf{81.58} \\
    Meta-Llama-3-8B-Instruct & 66.87 & 60.75 & 78.55 & 51.65 & 67.07 & 74.51 & 68.69 \\
    \textbf{Bielik-Minitron-7B-v3.0-Instruct} & \textbf{66.60} & \textbf{56.48} & \textbf{74.20} & \textbf{49.04} & \textbf{64.55} & \textbf{72.61} & \textbf{82.71} \\
    Mistral-7B-Instruct-v0.2 & 65.71 & 63.14 & 84.88 & 68.26 & 60.78 & 77.19 & 40.03 \\
    \underline{Bielik-4.5B-v3.0-Instruct} & \underline{64.89} & \underline{56.06} & \underline{73.90} & \underline{50.79} & \underline{63.66} & \underline{71.19} & \underline{73.69} \\
    gemma-7b & 64.29 & 61.09 & 82.47 & 44.91 & 66.03 & 78.45 & 52.77 \\
    Qwen1.5-32B-Chat & 62.95 & 66.04 & 85.49 & 66.95 & 74.99 & 77.19 & 7.05 \\
    Qwen1.5-14B-Chat & 62.27 & 58.70 & 82.27 & 60.36 & 68.57 & 73.09 & 30.63 \\
    Qwen1.5-7B-Chat & 55.15 & 55.89 & 78.56 & 53.54 & 61.65 & 67.72 & 13.57 \\
    Mistral-7B-Instruct-v0.1 & 54.96 & 54.52 & 75.63 & 56.28 & 55.38 & 73.72 & 14.25 \\
    \underline{Bielik-7B-Instruct-v0.1} & \underline{51.26} & 47.53 & 68.91 & 46.18 & 49.47 & 65.51 & 29.95 \\
    \bottomrule
    \end{tabular}
    \caption{Open LLM Leaderboard results for selected instruction-tuned models}
    \label{tab:open-llm-instruct}
    \end{table*}
\end{landscape}

On English language tasks (Table~\ref{tab:open-llm-instruct}), Bielik models demonstrate solid cross-lingual capabilities. Bielik-11B-v3.0-Instruct scores 72.45 average, with particularly strong performance on GSM8K (85.60) and ARC challenge (64.59), indicating robust mathematical and reasoning capabilities. The Polish tokenizer variants achieve 71.49 (Bielik-PL-11B-v3.0-Instruct) and 67.63 (Bielik-PL-Minitron-7B-v3.0-Instruct) on the same Open LLM Leaderboard aggregate.

\subsection{INCLUDE-base-44}

INCLUDE is a comprehensive knowledge- and reasoning-centric benchmark designed to evaluate multilingual language models across 44 languages in realistic deployment scenarios. The benchmark comprises 22,637 four-option multiple-choice questions extracted from academic and professional examinations, covering 57 topics across diverse domains including STEM (Biology, Chemistry, Physics, Mathematics, Computer Science), Arts \& Humanities (History, Philosophy, Literature, Visual Arts, Law), Social Sciences (Sociology, Economics, Psychology), Health-oriented Education (Medicine), and professional certifications (driving licenses, medical licenses, professional certifications).

A distinguishing feature of INCLUDE is its emphasis on regional knowledge and cultural context. Questions are categorized as either "agnostic" (universally applicable) or "region implicit/explicit" (requiring cultural or geographical knowledge specific to particular regions). This design enables assessment of models' ability to handle not only universal knowledge but also culturally-specific content essential for deployment in diverse linguistic communities. For our evaluation, we focus on a subset of 20 European languages from the full benchmark to assess Bielik's performance across its target linguistic region. The benchmark evaluation is presented in Table~\ref{tab:include-base-44}.

\begin{table*}[t]
  \centering
  \small
  \begin{tabular}{lrrr}
  \toprule
  \textbf{Model} & \textbf{Params (B)} & \textbf{AVG} & \textbf{Polish} \\
  \midrule
  \textbf{Bielik-11B-v3.0-Instruct} & \textbf{11.2} & \textbf{64.8} & \textbf{69.0} \\
  Qwen2.5-14B-Instruct & 14.8 & 61.7 & 58.9 \\
  \textbf{Bielik-11B-v3-Base-20250730} & \textbf{11.2} & \textbf{60.6} & \textbf{63.9} \\
  phi-4 & 14.7 & 58.8 & 49.6 \\
  Apertus-8B-Instruct-2509 & 8.0 & 57.9 & 49.6 \\
  \textbf{Bielik-Minitron-7B-v3.0-Instruct} & \textbf{7.5} & \textbf{57.4} & \textbf{59.3} \\
  Llama-3.1-8B-Instruct & 8.0 & 55.3 & 53.8 \\
  EuroLLM-9B-Instruct & 9.2 & 55.1 & 52.0 \\
  Qwen2.5-7B-Instruct & 7.6 & 54.4 & 52.2 \\
  \textbf{Bielik-PL-11B-v3.0-Instruct} & \textbf{11.2} & \textbf{53.92} & \textbf{64.23} \\
  Mistral-Nemo-Instruct-2407 & 12.2 & 53.2 & 48.4 \\
  \underline{Bielik-11B-v2.6-Instruct} & \underline{11.2} & \underline{51.5} & \underline{59.3} \\
  Mistral-Nemo-Base-2407 & 12.2 & 51.2 & 44.9 \\
  \textbf{Bielik-PL-Minitron-7B-v3.0-Instruct} & \textbf{7.5} & \textbf{49.81} & \textbf{59.49} \\
  EuroLLM-9B & 9.2 & 49.2 & 45.6 \\
  aya-expanse-8b & 8.0 & 45.3 & 46.4 \\
  Mistral-7B-Instruct-v0.2 & 7.0 & 45.3 & 44.7 \\
  \underline{Bielik-11B-v2} & \underline{11.2} & \underline{44.8} & \underline{53.5} \\
  pllum-12b-nc-chat-250715 & 12.0 & 44.2 & 60.6 \\
  Mistral-7B-v0.2 & 7.0 & 41.8 & 37.2 \\
  pllum-12b-nc-base-250715 & 12.0 & 37.8 & 52.7 \\
  \underline{Bielik-4.5B-v3} & \underline{4.8} & \underline{35.9} & \underline{48.7} \\
  PLLuM-12B-base-250801 & 12.0 & 35.5 & 44.5 \\
  Llama-PLLuM-8B-base-250801 & 8.0 & 30.0 & 37.2 \\
  \bottomrule
  \end{tabular}
  \caption{INCLUDE-base-44 benchmark results showing average performance across European languages (20 language subset) and Polish-specific scores.}
  \label{tab:include-base-44}
\end{table*}

On INCLUDE-base-44 (Table~\ref{tab:include-base-44}), Bielik-11B-v3.0-Instruct achieves the highest scores among the models listed, with 64.8 average across European languages and 69.0 on Polish-specific tasks. This demonstrates superior balanced multilingual performance within this comparison, surpassing Qwen2.5-14B-Instruct (61.7 average, 58.9 Polish) despite having fewer parameters. Notably, Bielik's Polish-specific score (69.0) substantially exceeds its multilingual average (64.8), reflecting the model's particular strength in its primary target language while maintaining robust cross-lingual capabilities. The base model Bielik-11B-v3 also shows strong performance (60.6 average, 63.9 Polish), outperforming several instruction-tuned models in the same table including Llama-3.1-8B-Instruct and EuroLLM-9B-Instruct. \textbf{Bielik-Minitron-7B-v3.0-Instruct} achieves 57.4 average and 59.3 on Polish \cite{kinas2026bielikminitron7bcompressinglargelanguage}. The Polish tokenizer variants, Bielik-PL-11B-v3.0-Instruct and Bielik-PL-Minitron-7B-v3.0-Instruct, reach 53.92 and 49.81 on the European-language average and 64.23 and 59.49 on Polish-specific tasks, respectively. Compared to the previous version, Bielik-11B-v2 achieved 44.8 average with 53.5 on Polish tasks, showing significant improvement in v3.

\subsection{Belebele Reading Comprehension}

Belebele \cite{bandarkar-etal-2024-belebele} is a massively multilingual reading comprehension benchmark spanning 122 language variants. The benchmark consists of multiple-choice reading comprehension questions derived from the FLORES-200 dataset, where models must demonstrate understanding of short passages by correctly answering questions about their content. For our evaluation, we assess performance across 28 European language variants to evaluate Bielik's reading comprehension capabilities across its target linguistic region. Results are presented in Table~\ref{tab:belebele}.

\begin{table*}[t]
  \centering
  \small
  \begin{tabular}{lrrr}
  \toprule
  \textbf{Model} & \textbf{Params (B)} & \textbf{AVG} & \textbf{Polish} \\
  \midrule
  Qwen2.5-14B-Instruct & 14.8 & 85.91 & 87.56 \\
  \textbf{Bielik-11B-v3.0-Instruct} & \textbf{11.2} & \textbf{82.98} & 82.11 \\
  phi-4 & 14.7 & 81.71 & 83.00 \\
  \textbf{Bielik-Minitron-7B-v3.0-Instruct} & \textbf{7.5} & \textbf{78.03} & 78.22 \\
  \textbf{Bielik-PL-11B-v3.0-Instruct} & \textbf{11.2} & \textbf{77.41} & \textbf{81.22} \\
  Qwen2.5-7B & 7.6 & 74.60 & 79.00 \\
  \textbf{Bielik-PL-Minitron-7B-v3.0-Instruct} & \textbf{7.5} & \textbf{74.23} & \textbf{77.44} \\
  Mistral-Nemo-Instruct-2407 & 12.2 & 74.14 & 71.44 \\
  cjvt/GaMS-9B-Instruct & 9.2 & 72.40 & 71.89 \\
  Apertus-8B-Instruct-2509 & 8.0 & 69.58 & 70.00 \\
  EuroLLM-9B-Instruct & 9.2 & 69.05 & 71.22 \\
  \underline{Bielik-11B-v2.6-Instruct} & \underline{11.2} & \underline{68.67} & 79.22 \\
  Apertus-8B-2509 & 8.0 & 59.04 & 58.44 \\
  \bottomrule
  \end{tabular}
  \caption{Belebele benchmark results: European-language average (28-language subset) and Polish-specific accuracy.}
  \label{tab:belebele}
\end{table*}

On Belebele (Table~\ref{tab:belebele}), Bielik-11B-v3.0-Instruct achieves 82.98 average across European languages, representing a substantial improvement over the previous version Bielik-11B-v2.6-Instruct (68.67). On this 28-language European subset average, this score places Bielik second among the models listed, closely following Qwen2.5-14B-Instruct (85.91) and ahead of the phi-4 14.7B model (81.71). \textbf{Bielik-Minitron-7B-v3.0-Instruct} reaches 78.03 on the European-language subset \cite{kinas2026bielikminitron7bcompressinglargelanguage}. The Polish tokenizer variants, Bielik-PL-11B-v3.0-Instruct and Bielik-PL-Minitron-7B-v3.0-Instruct, reach 77.41 and 74.23 on the European-language average and 81.22 and 77.44 on Polish-specific tasks, respectively, indicating a trade-off where the Polish-optimized checkpoints retain strong Polish accuracy while scoring lower on the multilingual European average.

\subsection{FLORES Machine Translation}

FLORES (FLORES-200) is a widely-used machine translation benchmark covering 200 languages, designed to evaluate translation quality across diverse linguistic families. The benchmark measures translation performance using BLEU scores, which assess n-gram overlap between model-generated translations and human reference translations. For Bielik evaluation, we assess translation performance across 20 European language pairs, focusing on bidirectional translations between Polish and other European languages, as well as translations among European languages. This evaluation provides insights into the model's multilingual translation capabilities across its target linguistic region. Results are shown in Table~\ref{tab:flores}.

\begin{table*}[t]
  \centering
  \small
  \begin{tabular}{lrrrr}
  \toprule
  \textbf{Model} & \textbf{Params (B)} & \textbf{AVG} & \textbf{to Polish} & \textbf{from Polish} \\
  \midrule
  EuroLLM-9B-Instruct$^{*}$ & 9.2 & 20.61 & 19.28 & 21.95 \\
  \textbf{Bielik-11B-v3.0-Instruct} & \textbf{11.2} & \textbf{19.22} & \textbf{18.54} & \textbf{19.91} \\
  \textbf{Bielik-11B-v3-Base-20250730} & \textbf{11.2} & \textbf{17.85} & \textbf{17.60} & \textbf{18.11} \\
  \textbf{Bielik-PL-11B-v3.0-Instruct} & \textbf{11.2} & \textbf{17.82} & \textbf{17.58} & \textbf{18.07} \\
  phi-4 (15B) & 14.7 & 15.58 & 14.55 & 16.61 \\
  \textbf{Bielik-Minitron-7B-v3.0-Instruct} & \textbf{7.5} & \textbf{15.53} & \textbf{15.74} & \textbf{15.32} \\
  \textbf{Bielik-PL-Minitron-7B-v3.0-Instruct} & \textbf{7.5} & \textbf{15.15} & \textbf{15.99} & \textbf{14.31} \\
  Mistral-Nemo-Instruct-2407 & 12.2 & 14.35 & 13.37 & 15.33 \\
  \underline{Bielik-11B-v2.6-Instruct} & \underline{11.2} & \underline{13.58} & \underline{15.77} & \underline{11.38} \\
  Qwen2.5-14B-Instruct & 14.8 & 13.24 & 12.55 & 13.93 \\
  Qwen2.5-7B-Instruct & 7.6 & 11.34 & 10.43 & 12.26 \\
  \underline{Bielik-11B-v2} & \underline{11.2} & \underline{11.25} & \underline{14.86} & \underline{7.64} \\
  \bottomrule
  \multicolumn{5}{l}{$^{*}$ EuroLLM was trained on FLORES dataset} \\
  \end{tabular}
  \caption{FLORES machine translation benchmark results showing translation performance across European languages (20 language pairs) measured by BLEU scores.}
  \label{tab:flores}
\end{table*}

On FLORES translation tasks (Table~\ref{tab:flores}), Bielik-11B-v3.0-Instruct achieves an average BLEU score of 19.22 across European language pairs, ranking second only to EuroLLM-9B-Instruct (20.61), which was trained on FLORES data. Notably, Bielik demonstrates balanced bidirectional translation capabilities with 18.54 BLEU for translation to Polish and 19.91 for translation from Polish. This represents a significant improvement over Bielik-11B-v2.6-Instruct (13.58 average), particularly in the from-Polish direction where v3.0 achieves 19.91 versus v2.6's 11.38. The base model Bielik-11B-v3 also shows strong translation performance (17.85 average), substantially outperforming larger models like phi-4 14.7B (15.58) and Qwen2.5-14B-Instruct (13.24). The Polish tokenizer checkpoints, \textbf{Bielik-PL-11B-v3.0-Instruct} and \textbf{Bielik-PL-Minitron-7B-v3.0-Instruct}, achieve 17.82 and 15.15 average BLEU (17.58/18.07 and 15.99/14.31 for to-Polish/from-Polish, respectively). \textbf{Bielik-Minitron-7B-v3.0-Instruct} (original tokenizer) achieves 15.53 average BLEU (15.74 to Polish, 15.32 from Polish) \cite{kinas2026bielikminitron7bcompressinglargelanguage}.

\subsection{Summary of Evaluation Results}

The Bielik 11B v3 family (original tokenizer) achieves strong results across Polish and multilingual benchmarks, as detailed above and in \cite{ociepa2025bielik11bv3multilingual}. For example, Bielik-11B-v3.0-Instruct reaches 65.93 on the Open PL LLM Leaderboard (5-shot), 71.20 on Polish EQ-Bench, and 72.45 average on the English Open LLM Leaderboard, among others (including 71.83\% on the Polish Linguistic and Cultural Competency Benchmark in the reference report).

The Bielik v3 PL models match this picture where evaluated. Table~\ref{tab:open-pl-llm-instruct} lists 5-shot Open PL LLM averages (65.93, 64.11, and 61.66 for Bielik-11B-v3.0-Instruct, Bielik-PL-11B-v3.0-Instruct, and Bielik-PL-Minitron-7B-v3.0-Instruct, respectively), with \textbf{Bielik-Minitron-7B-v3.0-Instruct} at 62.46 for comparison \cite{kinas2026bielikminitron7bcompressinglargelanguage}. Polish EQ-Bench (\texttt{eq-bench\_v2\_pl}) scores are 71.15 and 66.89 for the 11B and 7B PL variants, and the Polish Medical Leaderboard scores are 48.42\% and 43.35\%. The English Open LLM Leaderboard averages are 71.49 and 67.63 (Table~\ref{tab:open-llm-instruct}).

CPTUB overall averages for the PL checkpoints are 3.80 (11B) and 3.55 (7B Minitron). FLORES BLEU for the PL checkpoints is 17.82 average (17.58 to Polish, 18.07 from Polish) for the 11B model and 15.15 average (15.99 to Polish, 14.31 from Polish) for the 7B Minitron variant, with \textbf{Bielik-Minitron-7B-v3.0-Instruct} (original tokenizer) FLORES figures from \cite{kinas2026bielikminitron7bcompressinglargelanguage}. On INCLUDE-base-44 the PL checkpoints score 53.92/64.23 (11B) and 49.81/59.49 (7B Minitron) for European average and Polish, respectively, with the original-tokenizer Minitron at 57.4/59.3. On Belebele they score 77.41/81.22 (11B) and 74.23/77.44 (7B Minitron), with the original-tokenizer Minitron at 78.03 (European average). PLCC scores for the PL checkpoints are likewise pending.
\section{Limitations and Biases}

While the Bielik v3 PL models represent a state-of-the-art advancement for the Polish language, they possess standard LLM limitations. Models can produce factually incorrect output, and should not be relied on to produce factually accurate data. While great efforts have been taken to clear the training data, it is possible that this model can generate lewd, false, biased or otherwise offensive outputs.

\section{Conclusion}

In this technical report, we presented the Bielik v3 PL series - 11B and 7B parameter variants whose Mistral-derived tokenizer has been replaced with the Polish-optimized APT4 tokenizer. Despite keeping a comparable vocabulary size ($\sim$32{,}000 tokens), this change reduces the fertility ratio from 3.22 to 1.62 tokens per word on representative Polish text (Table~\ref{tab:tokenizers-comparison}), nearly doubling the effective Polish context capacity.

To mitigate catastrophic forgetting during vocabulary adaptation, we combined FOCUS-based embedding initialization with a two-stage continued pretraining pipeline (4B tokens with partial freezing, followed by 16B tokens of full adaptation) and applied the same post-training alignment (SFT, DPO-P, GRPO) as the original Bielik v3 models. Evaluation across nine Polish and multilingual benchmarks (Section~\ref{sec:evaluation}) confirms that the Bielik v3 PL models closely preserve - and on CPTUB and Polish EQ-Bench even surpass - the performance of their original-tokenizer counterparts, while English-language capabilities remain largely intact.

Both models are released under the Apache 2.0 license. The methodology described here - FOCUS-based vocabulary transfer, staged pretraining with progressive unfreezing, and consistent post-training alignment - provides a reproducible blueprint for adapting multilingual large language models to specific languages with improved tokenization efficiency.

\section*{Acknowledgements}

We gratefully acknowledge Polish high-performance computing infrastructure PLGrid (HPC Center: ACK Cyfronet AGH) for providing computer facilities and support within computational grant no. PLG/2024/016951. 

The model could not have been created without the commitment and work of the entire SpeakLeash team, whose contribution is invaluable. Thanks to the hard work of many individuals, it was possible to gather a large amount of content in Polish and establish collaboration between the open-science SpeakLeash project and the HPC center: ACK Cyfronet AGH. Individuals who contributed to the creation of the model through their commitment to the open-science SpeakLeash project: Sebastian Kondracki, Marek Magry\'{s}, Igor Ciuciura, Dominika Basaj, Kuba So\l{}tys, Karol Jezierski, Sonia Staniek, Anna Przyby\l{}, and many other wonderful researchers and enthusiasts of the AI world.

\bibliographystyle{unsrtnat}
\bibliography{biblio}

\clearpage

\appendix

\section{The preamble of the Constitution of the Republic of Poland} \label{app:preamble}

\paragraph{Polish}

W trosce o byt i przyszłość naszej Ojczyzny,

odzyskawszy w 1989 roku możliwość suwerennego i demokratycznego stanowienia o Jej losie,

my, Naród Polski - wszyscy obywatele Rzeczypospolitej,

zarówno wierzący w Boga będącego źródłem prawdy, sprawiedliwości, dobra i piękna,

jak i nie podzielający tej wiary, a te uniwersalne wartości wywodzący z innych źródeł,

równi w prawach i w powinnościach wobec dobra wspólnego - Polski,

wdzięczni naszym przodkom za ich pracę, za walkę o niepodległość okupioną ogromnymi ofiarami, za kulturę zakorzenioną w chrześcijańskim dziedzictwie Narodu i ogólnoludzkich wartościach,

nawiązując do najlepszych tradycji Pierwszej i Drugiej Rzeczypospolitej,

zobowiązani, by przekazać przyszłym pokoleniom wszystko, co cenne z ponad tysiącletniego dorobku,

złączeni więzami wspólnoty z naszymi rodakami rozsianymi po świecie,

świadomi potrzeby współpracy ze wszystkimi krajami dla dobra Rodziny Ludzkiej,

pomni gorzkich doświadczeń z czasów, gdy podstawowe wolności i prawa człowieka były w naszej Ojczyźnie łamane,

pragnąc na zawsze zagwarantować prawa obywatelskie, a działaniu instytucji publicznych zapewnić rzetelność i sprawność,

w poczuciu odpowiedzialności przed Bogiem lub przed własnym sumieniem,

ustanawiamy Konstytucję Rzeczypospolitej Polskiej jako prawa podstawowe dla państwa oparte na poszanowaniu wolności i sprawiedliwości, współdziałaniu władz, dialogu społecznym oraz na zasadzie pomocniczości umacniającej uprawnienia obywateli i ich wspólnot.

Wszystkich, którzy dla dobra Trzeciej Rzeczypospolitej tę Konstytucję będą stosowali, wzywamy, aby czynili to, dbając o zachowanie przyrodzonej godności człowieka, jego prawa do wolności i obowiązku solidarności z innymi, a poszanowanie tych zasad mieli za niewzruszoną podstawę Rzeczypospolitej Polskiej.

\paragraph{English}

Having regard for the existence and future of our Homeland,

Which recovered, in 1989, the possibility of a sovereign and democratic determination of its fate,

We, the Polish Nation - all citizens of the Republic,

Both those who believe in God as the source of truth, justice, good and beauty,

As well as those not sharing such faith but respecting those universal values as arising from other sources,

Equal in rights and obligations towards the common good - Poland,

Beholden to our ancestors for their labours, their struggle for independence achieved at great sacrifice, for our culture rooted in the Christian heritage of the Nation and in universal human values,

Recalling the best traditions of the First and the Second Republic,

Obliged to bequeath to future generations all that is valuable from our over one thousand years' heritage,

Bound in community with our compatriots dispersed throughout the world,

Aware of the need for cooperation with all countries for the good of the Human Family,

Mindful of the bitter experiences of the times when fundamental freedoms and human rights were violated in our Homeland,

Desiring to guarantee the rights of the citizens for all time, and to ensure diligence and efficiency in the work of public bodies,

Recognizing our responsibility before God or our own consciences,

Hereby establish this Constitution of the Republic of Poland as the basic law for the State, based on respect for freedom and justice, cooperation between the public powers, social dialogue as well as on the principle of subsidiarity in the strengthening the powers of citizens and their communities.

We call upon all those who will apply this Constitution for the good of the Third Republic to do so paying respect to the inherent dignity of the person, his or her right to freedom, the obligation of solidarity with others, and respect for these principles as the unshakeable foundation of the Republic of Poland.

\end{document}